\documentclass[sigconf]{aamas} 
\usepackage{balance} % for balancing columns on the final page
\usepackage{hyperref}
\setcopyright{none}
\acmConference[AAMAS '22]{Proc.\@ of the 21st International Conference
on Autonomous Agents and Multiagent Systems (AAMAS 2022)}{May 9--13, 2022}
{Online}{ P. Faliszewski, V. Mascardi, C. Pelachaud, M.E. Taylor (eds.)}
\copyrightyear{2022}
\acmYear{2022}
\acmDOI{}
\acmPrice{}
\acmISBN{}

\setcopyright{none}
\renewcommand\footnotetextcopyrightpermission[1]{}
\acmSubmissionID{9}

\title{JEDAI: A System for Skill-Aligned Explainable Robot Planning}
\author{Naman Shah}
\authornote{Equal Contribution. Alphabetical Order.}
\affiliation{
  \institution{Arizona State University, USA}
  }
\email{shah.naman@asu.edu}

\author{Pulkit Verma}
\authornotemark[1]
\affiliation{
  \institution{Arizona State University, USA}
  }
\email{verma.pulkit@asu.edu}

\author{Trevor Angle}
\affiliation{
  \institution{Arizona State University, USA}
  }
\email{taangle@asu.edu}

\author{Siddharth Srivastava}
\affiliation{
  \institution{Arizona State University, USA}
  }
\email{siddharths@asu.edu}

\begin{abstract}
This paper presents JEDAI,
an AI system designed for outreach and educational efforts
aimed at non-AI experts.
JEDAI features a novel synthesis of research ideas from
integrated task and motion planning and explainable AI.
JEDAI helps users create high-level, intuitive plans
while ensuring that they will be executable by the robot.
It also provides users customized explanations about errors
and helps improve their understanding of AI planning
as well as the limits and capabilities of the underlying robot system.
\end{abstract}

\keywords{AI in Education; Explanations; Task and Motion Planning; Robotics}

\newcommand{\BibTeX}{\rm B\kern-.05em{\sc i\kern-.025em b}\kern-.08em\TeX}
\newcommand{\mysssection}[1]{\noindent\textbf{#1}\hspace{4pt}}
\newcommand{\pvsection}[1]{\vspace{0.03in}\mysssection{#1}}

\newcommand\blfootnote[1]{%
  \begingroup
  \renewcommand\thefootnote{}\footnote{#1}%
  \addtocounter{footnote}{-1}%
  \endgroup
}

\begin{document}

%%% The following commands remove the headers in your paper. For final 
%%% papers, these will be inserted during the pagination process.

\pagestyle{fancy}
\fancyhead{}

%%% The next command prints the information defined in the preamble.

\maketitle 

%%%%%%%%%%%%%%%%%%%%%%%%%%%%%%%%%%%%%%%%%%%%%%%%%%%%%%%%%%%%%%%%%%%%%%%%

\section{Introduction}
\label{sec:introduction}
\blfootnote{This paper is originally published in Proc. of 21st Intl. Conference on Autonomous Agents and Multiagent Systems (AAMAS, 2022). Please cite the original work as: \\ Naman Shah, Pulkit Verma, Trevor Angle, and Siddharth Srivastava. 2022. JEDAI: A System for Skill-Aligned Explainable Robot Planning: Demonstration Track. In Proc. of the 21st International Conference on Autonomous Agents
and Multiagent Systems (AAMAS 2022), Online, May 9–13, 2022, IFAAMAS,
3 pages.}

AI systems are increasingly common in everyday life,
where they can be used by laypersons who may not
understand how these autonomous systems work or what they can
and cannot do.
This problem is particularly salient in cases of
taskable AI systems whose functionality can change
based on the tasks they are performing.
In this work, we present an AI system JEDAI
(JEDAI Explains Decision-Making AI) that can be used
in outreach and educational efforts to help laypersons learn how to
provide AI systems with new tasks, debug such systems, and understand
their capabilities.

The research ideas brought together in JEDAI address
three key technical challenges:
(i) abstracting a robot's functionalities into
high-level actions (capabilities) that the user can more easily understand;
(ii) converting the user-understandable
capabilities into low-level motion plans that
a robot can execute; and (iii) explaining errors in a
manner sensitive to the user's current level of knowledge
so as to make the robot's capabilities and limitations clear.

\begin{figure}
    \centering
    \includegraphics[width=\columnwidth]{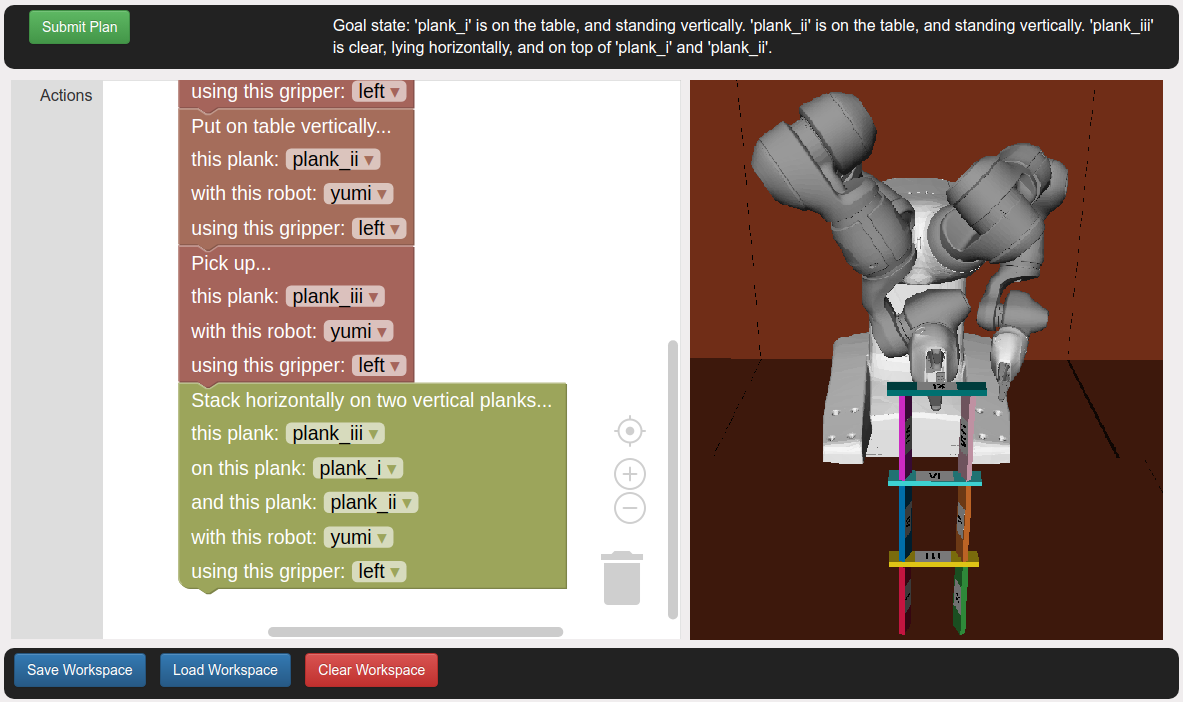}
    \caption{JEDAI system with a Blockly-based plan creator
    on the left and a simulator window on the right.}
    \label{fig:jedai}
    \Description{ The image shows the GUI interface of the presented framework. On the left it shows a partial plan created by the user using Blockly coding interface and on the right, it shows an image of the goal.
    }
\end{figure}

JEDAI utilizes recent work in explainable AI and integrated task and motion planning
to address these challenges and provides a simple 
interface to support accessibility.
Users select a domain and an associated task,
after which they create a plan
consisting of high-level actions (Fig.~\ref{fig:jedai} left)
to complete the task.
The user puts together a plan in a drag-and-drop workspace,
built with the Blockly visual programming library~\cite{google_blockly}.
JEDAI validates this plan using the Hierarchical Expertise Level Modeling
algorithm (HELM)~\cite{sreedharan_2018_helm,sreedharan_2021_helm}.
If the plan contains any errors, HELM computes a
user-specific explanation of why the plan would fail.
JEDAI converts such explanations to natural language,
thus helping to identify and fix
any gaps in the user's understanding.
Whereas, if the plan given by the user is a correct
solution to the current task, JEDAI uses a task and motion planner
ATM-MDP~\cite{shah_2020_anytime,shah_2021_anytime} to
convert the high-level plan,
that the user understands, to a low-level 
motion plan that the robot can execute.
The user is shown the execution of this low-level motion plan by the 
robot in a simulated environment (Fig.~\ref{fig:jedai} right).

Prior work 
on the topic includes approaches that solve the three
technical challenges mentioned earlier
\textit{in isolation}. This includes
tools for: providing
visualizations or animations of standard planning 
domains~\cite{magnaguagno_2017_web,chen2020planimation,Aguinaldo_2021_graphical,dvorak_2021_visual,dePellegrin_2021_pdsim,Roberts_2021_vplansim};
making it easier for non-expert users to program robots
with low-level actions~\cite{Krishnamoorthy_2016_using,
Weintrop_2018_evaluating,huang_2020_vipo,winterer_2020_expert};
and generating explanations for plans provided by the 
users~\cite{grover_2020_radar,Karthik_2021_radarx,brandao2021towards,kumar2021vizxp}.
In addition, none of these works make the instructions easier for the user,
have the ability to automatically compute user-aligned explanations, and work with real robots (or their simulators) at the same time. JEDAI addresses all three challenges in tandem by using 3D simulations for domains with real robots and their actual constraints and providing personalized explanations that inform a user of any mistake they make while using the system.

\section{Architecture}

\begin{figure}[t!]
    \centering
    \includegraphics[width=\columnwidth]{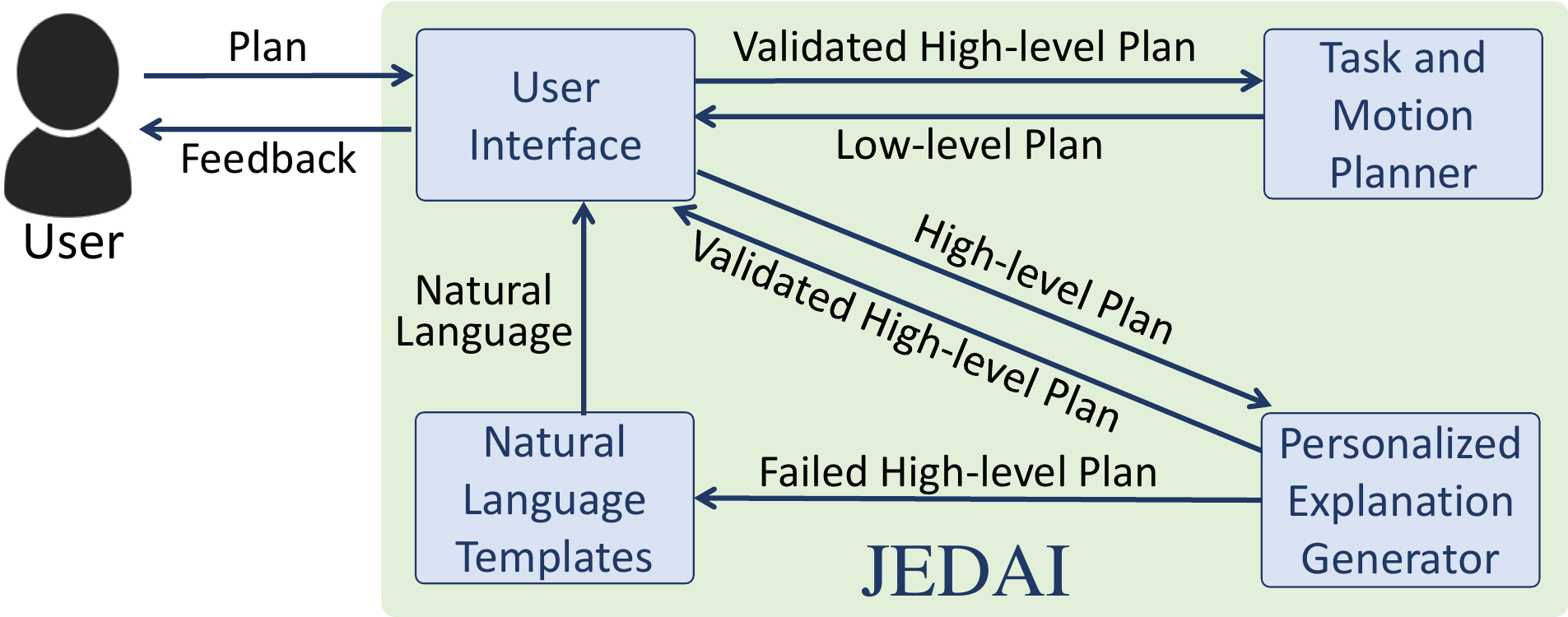}
    \caption{Architecture of JEDAI showing interaction between the four core components.}
    \label{fig:system}
    \Description{The image shows the overall architecture of the JEDAI system. It shows connections between four core components of the architecture: user interface, task and motion planner, natural language templates, and personalized explanation generator. It also shows how they communicate with each other and with the end user.}
\end{figure}

Fig. \ref{fig:system} shows the four core components of the JEDAI framework: 
(i)~user interface, (ii)~task and motion planner, 
(iii)~personalized explanation generator, and (iv)~natural language templates. 
We now describe each component in detail.

\pvsection{User interface}
JEDAI's user interface (Fig.~\ref{fig:jedai}) is made to be unintimidating and easy to use.
The Blockly visual programming interface is used
to facilitate this. 
JEDAI generates a separate interconnecting block for each high-level action,
and action parameters are picked from drop-down selection fields that display
type-consistent options for each parameter.
Users can drag-and-drop these actions and select different arguments
to create a high-level plan.

\pvsection{Personalized explanation generator}
Users will sometimes make mistakes when planning,
either failing to achieve goal conditions or applying actions
before the necessary preconditions are satisfied.
For inexperienced users in particular,
these mistakes may stem from an incomplete understanding
of the task’s requirements or the robot’s capabilities.
JEDAI assists users in apprehending these details
by providing explanations personalized to each user.

Explanations in the context of this work are of two types:
(i) non-achieved goal conditions, and
(ii) violation of a precondition of an action.
JEDAI validates the plan submitted by the user to check if
it achieves all goal conditions. 
If it fails to achieve
any goal condition, the user is informed about it.
JEDAI uses HELM to compute user-specific contrastive explanations
in order to explain any unmet precondition in an action
used in the user's plan.
HELM does this by using the plan
submitted by the user to estimate the user's understanding of the
robot's model
and then uses the estimated model to compute the personalized
explanations. 
In case of multiple errors in the user's plan, HELM generates explanation for one of the errors. This is
because explaining the reason for more than one errors might be unnecessary and in the worst case might leave the user feeling overwhelmed~\cite{Miller_2019_explanation}. 
An error is selected for explanation by HELM based on optimizing a cost function that indicates the relative difficulty of concept understandability which can be 
changed to reflect different users' background knowledge.

\pvsection{Natural language templates}
Even with a user-friendly interface and
personalized explanations for errors in
abstract plans, domain model syntax
used for interaction with ATM-MDP
presents a significant barrier
to a non-expert trying to understand
the state of an environment
and the capabilities of a robot. To alleviate this,
JEDAI uses language templates that use the structure of the planning formalism for generating natural language descriptions for goals, actions, and explanations. E.g., the action \emph{``pickup (plank\_i gripper\_left)''} can be described in 
natural language as ``pick up \emph{plank\_i} with \emph{the left gripper}''.
Currently, we use hand-written templates for these translations, but an automated approach can also be used.

\pvsection{Task and motion planner}
JEDAI uses ATM-MDP to convert the high-level plan submitted by the user
into sequences of low-level primitive actions that a robot can execute.

ATM-MDP uses sampling-based motion planners
to provide a probabilistically complete approach to hierarchical planning.
High-level plans are refined by
computing feasible motion plans for each high-level action.
If an action does not accept any valid refinement
due to discrepancies between the symbolic state
and the low-level environment, it reports the failure back to JEDAI.
If all actions in the high-level plan are refined successfully,
the plan's execution is shown using the OpenRAVE simulator~\cite{Diankov_2008_openrave}.

\pvsection{Implementation}
Any custom domain can be set up with JEDAI. We provide five built-in domains,
each with one of YuMi~\cite{yumi} or Fetch~\cite{wise16_fetch} robots.
Each domain contains a set of problems that the users can attempt to solve
and low-level environments corresponding to these problems. Source code for
the framework, an already setup virtual machine, and the documentation are
available at: 
\href{https://github.com/aair-lab/AAIR-JEDAI}{\small \texttt{https://github.com/aair-lab/AAIR-JEDAI}}.
A video demonstrating JEDAI's working is available at:
\href{https://youtu.be/MQdoikcnhbY}{\small \texttt{https://youtu.be/MQdoikcnhbY}}.

\section{Conclusions and Future Work}
We demonstrated a novel AI tool JEDAI for
helping people understand the capabilities
of an arbitrary AI system and enabling them
to work with such systems.
JEDAI converts the user's input plans to low level
motion plans executable by the robot if it is correct,
or explains to the user any error in the plan if it is incorrect.
JEDAI works with
off-the-shelf task and motion planners
and explanation generators. This structure allows it to scale automatically with
improvements in either of these active research areas. JEDAI's vizualization-based
interface could also be used to foster trust in AI systems~\cite{Beauxis21_role}.

JEDAI uses predefined abstractions to verify plans provided by the user. In the future, we plan on extending it to learn abstractions automatically~\citep{shah2022using}. JEDAI could also be extended as an interface for assessing an agent's functionalities and capabilities by interrogating the agent~\citep{verma_21_asking,nayyar2022differential,verma2022discovering} as well as to
work as an interface that makes AI systems 
compliant with Level II assistive AI – systems
that makes it easy for operators to learn how to use them
safely~\cite{srivastava_2021_unifying}. 
Extending this tool for 
working in 
non-stationary settings, 
and generating 
natural language
descriptions of
predicates and actions autonomously are a few other promising directions of
future work.

\section*{Acknowledgements}
We thank Kiran Prasad and Kyle Atkinson for help with the
implementation, Sarath Sreedharan for help
with setting up HELM, and Sydney Wallace for feedback on
user interface design. We also thank Chirav Dave, Rushang Karia,
Judith Rosenke, and Amruta Tapadiya for their work on
an earlier version of the system.
This work was supported in part by the NSF grants IIS 1909370,
IIS 1942856, IIS 1844325, OIA 1936997, and the ONR grant N00014-21-1-2045.

%%%%%%%%%%%%%%%%%%%%%%%%%%%%%%%%%%%%%%%%%%%%%%%%%%%%%%%%%%%%%%%%%%%%%%%%

%%% The next two lines define, first, the bibliography style to be 
%%% applied, and, second, the bibliography file to be used.

\bibliographystyle{ACM-Reference-Format} 
\balance
\bibliography{jedai}

%%%%%%%%%%%%%%%%%%%%%%%%%%%%%%%%%%%%%%%%%%%%%%%%%%%%%%%%%%%%%%%%%%%%%%%%

\end{document}